\documentclass{article}

\usepackage[dblblindworkshop, final]{neurips_2026}
\workshoptitle{Workshop for Autonomous ML Research}

\usepackage[utf8]{inputenc}
\usepackage[T1]{fontenc}
\usepackage{hyperref}
\usepackage{url}
\usepackage{booktabs}
\usepackage{amsfonts}
\usepackage{amssymb}
\usepackage{amsmath}
\usepackage{nicefrac}
\usepackage{microtype}
\usepackage{xcolor}
\usepackage{graphicx}

\title{MLE Agents: Autonomous ML Engineering in Large-scale Recommendation Systems}

\author{%
  Matthew Li\thanks{Equal contribution.} \\
  Meta Platforms \\
  \texttt{mli01@meta.com} \\
  \And
  Albert Jiongqian Liang\footnotemark[1] \\
  Meta Platforms \\
  \texttt{albertlg@meta.com} \\
  \And
  Dilip Raj \\
  Meta Platforms \\
  \texttt{dilipraj@meta.com}
}

\begin{document}

\maketitle

\begin{abstract}
The typical workflow for improving large-scale ranking and recommendation models is slow and tedious: an engineer forms a hypothesis, implements it as a code change, trains a model, and evaluates it both offline and online, often relying on their own intuition and experience along the way. Autonomous LLM agents promise to speed this up by running the loop on their own, but it is still unclear whether they can do this in industry-scale systems rather than on benchmarks or small proxy tasks. We build human-on-the-loop machine learning engineer (MLE) agents that propose, implement, train, and evaluate model changes on large-scale search rankers. Engineers retain control over objective design, search direction, and final-model review. Each agent writes a model change as code, launches a training job, reads the offline evaluation, and decides whether to keep it. We identify three obstacles that make this industry setting challenging: feedback takes hours per experiment, offline metrics do not always predict online impact, and improvements saturate after the early wins. We address these with parallel training runs, an objective calibrated to match online impact, and careful context engineering (a "what didn't work" blocklist, strict output formats, and an anti-staleness rule). In one campaign, the system ran 1,097 completed candidate experiments and accepted 325, a 29.6\% acceptance rate compared with 15.6\% in the earlier campaign under a different objective and search constraint. Its best model increases search volume by 0.23\% in an A/B test of large scale. Several of these improvements were architectural changes the team had never tried -- mechanisms drawn from the broader industry and literature that had not previously been considered for this stack -- suggesting that MLE agents can explore regions of the design space that human intuition overlooks. Notably, we find that the design of the harness around the model matters: intuition suggests that a stronger underlying model is enough, but in our experience, the design of the loop around it is just as critical. More broadly, our work shows that autonomous agents can do real machine-learning engineering in industry-scale systems.
\end{abstract}

\noindent\textbf{Keywords:} MLE agents, Autonomous ML, Recommendation Systems, Ranking
models, LLM agents

\begin{figure}[b]
  \centering
  \includegraphics[width=\linewidth]{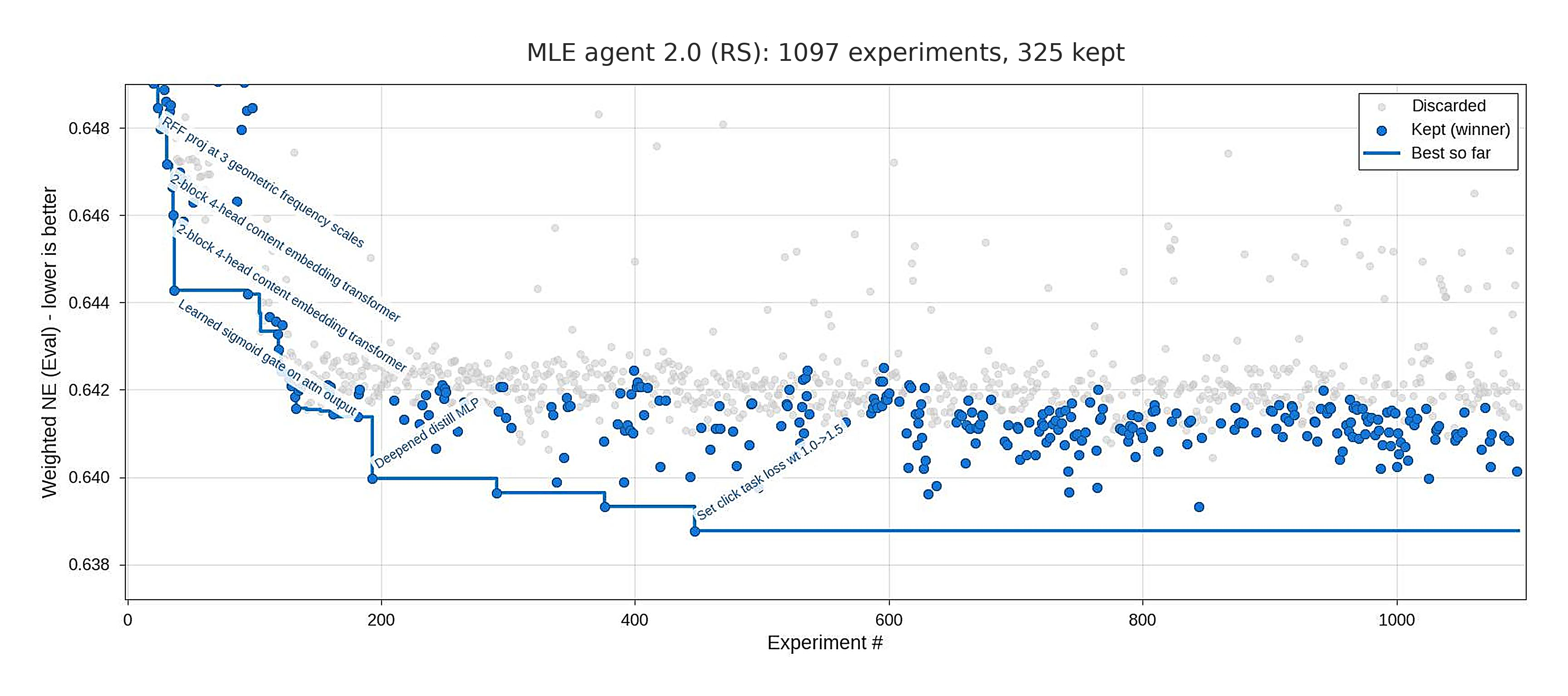}
  \caption{MLE agents 2.0 progress on Nullstate (weighted NE, lower is better):
  1{,}097 distinct configurations, 325 kept. Best-so-far staircase annotated with the
  discoveries that moved it; the curve flattens after the early wins.}
  \label{fig:progress}
\end{figure}

\section{Autonomous research result}
\label{sec:result}

\subsection{Hypothesis}
Can an autonomous MLE agent discover a sequence of model improvements for a large-scale search ranking system with over a billion requests per day that (a) compound into an offline quality improvement, (b) transfer reliably to online user impact, and (c) include genuinely new architectural contributions beyond what the human team had explored? A positive result requires an agent-discovered model that improves a calibrated offline objective and produces a statistically significant online improvement -- while surfacing at least some architectural changes the team had not previously attempted.

\subsection{Method}
The experiment was deliberately kept simple: let the agent run one long campaign against an objective baseline and incrementally keep what survives. We judge it on whether the final model with accumulated changes makes an online improvement. The target model is a multi-task ranking system that scores search query suggestions across multiple engagement objectives (click and various downstream actions). It serves a major search platform, using a transformer-based architecture with sparse and dense feature processing, multi-task prediction towers, and a distillation pathway.

\paragraph{Baseline.}
An large-scale ranker trained on multiple partitions of user-engagement data on the scale of hundreds-of-millions of examples and evaluated on held-out traffic. Baseline metrics: click NE = 0.7056 and content\_downstream NE = 0.6421 on the keyword + high-confidence evaluation slice.

\paragraph{Calibrated objective.}
A weighted composite of normalized entropy (NE; lower is better):
\begin{equation}
\mathcal{L} = 0.3 \times \text{click\_NE} + 0.7 \times \text{content\_downstream\_NE}.
\label{eq:objective}
\end{equation}
We chose the offline evaluation objective based on the alignment we observed between offline model metrics and online user engagement in earlier experiments. Because improvements in predicting downstream engagement were more closely aligned with online improvements than improvements in click prediction alone, we gave downstream engagement greater weight while retaining click prediction in the objective.
A candidate "wins" only if it improves the weighted objective beyond a variance-probe threshold of 0.0005 NE, ensuring improvements exceed training noise. We set this threshold using the average difference in weighted NE observed across historical variance-probe runs.

\paragraph{Search protocol.}
Both generations operate in an iterative loop: the agent proposes one coherent model modification per candidate, trains it on large-scale infrastructure, and records the evaluation and acceptance decision in persistent model-state logs. MLE agents 1.0 evaluated one candidate at a time using Click AUPRC; MLE agents 2.0 uses the calibrated objective of Equation~\ref{eq:objective} and evaluates nominally seven candidates plus a variance probe in parallel per iteration. Because each candidate takes hours to train and evaluate, parallel execution increases experiment throughput without making an individual training run faster.

\subsection{Evidence}

\paragraph{Offline metrics.}
We present findings from MLE agents 2.0 first since it was the most successful. Across 1,097 completed candidate runs over 131 iterations -- 1,252 completed runs in total, the remaining 155 being variance probes -- the agent accepted 325 candidates, a win rate of 29.6\% of candidate runs, converging to a cumulative improvement of -0.0224 \textbf{weighted NE} (weighted baseline
0.66110.6388). The improvement is driven more by content\_downstream (-0.0273 NE) than click (-0.0110 NE), consistent with the 0.7 content weight in
Equation~\ref{eq:objective}. Both task metrics improved in the final model, so the end-to-end result does not show a sacrifice in click prediction; it does not establish that click prediction had saturated. The primary offline result is a stack of distinct architectural and objective edits that compound; the largest contributors are shown in
Table~\ref{tab:discoveries}. Some other interesting structural wins that the agent tried include a backbone swap to a fused Self-Attention projection MLP with a 4-layer SE-DHEN (Squeeze-and-Excitation Deep and Hierarchical Ensemble Network) block, and bidirectional dense/sparse cross-attention with a dedicated content-embedding transformer.

\begin{table}[h]
  \caption{Largest contributors among the changes retained in the final model. Each contribution
  is marginal: the change in the weighted objective when that change was carried forward on top of
  those already retained. Contributions do not sum to the end-to-end result, since some retained
  steps moved the objective the wrong way.}
  \label{tab:discoveries}
  \centering
  \small
  \begin{tabular}{lc}
    \toprule
    Change (agent-generated) &  weighted NE \\
    \midrule
    content\_downstream neg\_label\_smoothing 0.020.005 & -0.0041 \\
    Remove content\_downstream neg\_label\_smoothing; content\_downstream loss\_weight 2.5 & -0.0038 \\
    Learned sigmoid gate on attention output (pre-FFN) in Self-Attention transformer blocks & -0.0024 \\
    Deeper attentive-distillation MLP [1024,512][1024,768,512], zero-init residual & -0.0021 \\
    Gated residual bypass from attentive-distillation output into task towers & -0.0021 \\
    Self-Attention transformer 43 blocks with FFN expansion 14 & -0.0020 \\
    content\_downstream loss\_weight 2.53.0 with reduced click neg\_label\_smoothing & -0.0020 \\
    Click loss\_weight 1.5 with dropout 0.08 on the click tower output & -0.0017 \\
    Random Fourier Feature projections replacing squared-feature gating in DenseArch & -0.0013 \\
    Compound: RFF gating with per-task K=4 sparse attention & -0.0012 \\
    \bottomrule
  \end{tabular}
\end{table}

\paragraph{Online metrics.}
To measure real-world impact, we ran a standard online A/B test - a control arm served by the existing ranker, and a test arm served by the new ranker carrying the MLE-agent-discovered optimizations. Differences between the two groups' behaviors across a wide range of engagement metrics are measured, from how often users search to whether they follow through to downstream actions. On both primary metrics the agent-discovered model performed statistically significantly better than the existing baseline, including against a refreshed model, with downstream engagement metrics directionally positive; the main metrics and their 95\% confidence intervals are listed in Table~\ref{tab:online}.

\begin{table}[h]
  \caption{{Online A/B Results. Measured over more than a week of exposure.}
}
  \label{tab:online}
  \centering
  \begin{tabular}{lc}
    \toprule
    Online metric & Impact \\
    \midrule
    Search session volume & $\mathbf{+0.23\%\pm0.15}$\\
    Downstream engagement (participation, retention) & Directionally positive\\
    \bottomrule
  \end{tabular}
\end{table}

\paragraph{Provenance of the online numbers.} The online results reported here are from the final experiment review, after the A/B test reached its full data read. The treatment arm served the consolidated MLE agents 2.0 checkpoint -- the same stack whose offline result is reported above -- so the offline and online results refer to one campaign and one model.

\paragraph{Online experiment details.} The test ran over one week. Reported movements are relative changes with 95\% confidence intervals computed by Fieller's method.

\paragraph{Guardrail regression.} The treatment also showed a latency regression on our surface time-to-render guardrail. We report it alongside the engagement improvements: the launch traded a measurable latency cost against the engagement improvement, and an agent optimizing a pure quality objective does not surface such trade-offs on its own.

\subsection{Primary Result}
At the end of the day, the offline improvements are good. More importantly, we observed the online improvement (+0.23\% search volume). Not only was that an improvement on online metrics, not a benchmark score, but several of the changes the agent found were architectural ideas the team had never tried, drawn from the broader industry and literature rather than from this team's own design history. The offline improvements still matter and are impressive, but mainly as a fast signal for the agent to iterate against. The success from these MLE agent runs encouraged us to invest further in this direction.

\section{Related Work}
Recent work has studied LLM agents for automated scientific discovery and
machine-learning engineering. The AI Scientist~\citep{aiscientist} and
AIDE~\citep{aide} use language models to formulate hypotheses, implement
experiments, and iteratively improve empirical results.
MLE-bench~\citep{mlebench} and RE-Bench~\citep{rebench} evaluate related
capabilities on curated ML-engineering and research tasks. These studies
establish that LLM agents can perform substantial experimental work, but most
of their evidence comes from benchmarks or offline task suites.

The closest industry-scale comparison is the Self-Evolving Recommendation
System of \citet{wang2026selfevolving}. Their system combines a high-frequency
offline optimization loop with a lower-frequency online deployment loop at
YouTube. It uses specialized optimizer, architecture, and reward personas, a
shared Experiment Journal, and a state machine for validating, training,
launching, and monitoring candidates. They report multiple statistically
significant launches, increased experiment throughput, and
controlled ablations showing the importance of the controller model, persona
prompting, and experiment-history organization.

Our work is complementary rather than a claim of the first large-scale
MLE-agent system. We provide independent evidence on a different
stack: a multi-task search-ranking model with sparse and dense features,
hours-long distributed training jobs, and an offline-to-online evaluation gap.
Our study emphasizes three aspects. First, we report a longitudinal campaign
with more than 1,000 candidate evaluations, exposing saturation and repeated
failure modes that are difficult to observe in shorter studies. Second, we
focus on reliability mechanisms for the offline loop, including parallel
variance probes, structured memory of failed hypotheses, strict output
contracts, and an anti-staleness policy. Third, we retain explicit human control over objective design, search redirection, final-model review, and online experiment, and document the operational challenges and lessons from deploying MLE agents in an industry-scale environment. 

\section{System Design and Meta-Analysis}
We now provide a meta-analysis of the research process.

\subsection{Agent and Harness}

We develop two generations of MLE agents: \textbf{MLE agents 1.0}, an exploratory version run on two industry-scale ranking models (Nullstate and Typeahead)\footnote{\emph{Nullstate} ranks the query suggestions shown when a user taps the search bar, before typing anything. \emph{Typeahead} ranks autocomplete suggestions as the user types.}, and \textbf{MLE agents 2.0}, a redesign informed by 1.0's limitations and run on Nullstate. Each agent receives structured context about the current model state, past experiment results, and strategic guidance, then generates code modifications to the model configuration, which are submitted as distributed training jobs on industry-scale infrastructure. MLE agents 1.0 evaluates one candidate at a time. In 2.0, each iteration begins with a Claude Opus 4.8 research call to gather architectural ideas, followed by a brainstorming call to select seven candidate changes. Seven implementation calls then run in parallel, one per candidate. The resulting model variants and a variance probe are trained in parallel, and completed runs are evaluated against the calibrated objective. Intuition suggests the lever is a stronger underlying model, and in our limited experience the design of the loop around it also makes a big difference. We never tested other models, so we cannot say how the two compare. What we can say is that an important driver of quality we observed was context engineering and objective calibration: the controller LLM was held fixed throughout, so every improvement reported in this paper came from harness and objective design rather than a stronger base model; the agent itself changed the ranker. The headline campaign in Section~\ref{sec:result} is a 2.0 run on Nullstate; Section~\ref{sec:loop} traces the path from 1.0 to 2.0.

\subsection{Tools}
The agent operates through an industry-scale ML platform. It edits the model's configuration and code directly under a strict output contract, then launches each candidate as a distributed training job on Grand Teton class trainers with 64 GPUs (8x8, batch size 256). Early 1.0 experiments typically required approximately 6--10 hours for training and evaluation. Following training-throughput optimizations, including less frequent metrics reporting, completed candidate and variance-probe runs in the reported 2.0 campaign had a median recorded elapsed time of 3.54 hours and a mean of 4.02 hours. These timings include submission, scheduling, training, evaluation, and metric-retrieval delays, rather than active training time alone. The campaigns were not a matched runtime benchmark, so the reduction should not be attributed solely to throughput tuning. When a job finishes, the agent reads the calibrated offline metrics -- click and content normalized entropy on both the full and the keyword+high-confidence slices -- and compares them against the current best. Campaign state (every hypothesis, commit hash, job ID, metric, and accept/reject decision) persists in a json state file, backed by a seeds file and an automatic path that resubmits jobs after transient infrastructure failures; a spreadsheet mirrors it for monitoring. Finally, in a dedicated research phase the agent searches the model codebase and the external literature (e.g., FRNet, BilinearSE, MaskNet, FiLM, differential attention) for architectural ideas worth trying.

\paragraph{Compute.} Across the 2.0 campaign, 1{,}252 runs completed (1{,}097 candidate runs and 155 variance probes), with a median recorded elapsed time of 3.54 hours (mean 4.02; 10th--90th percentile 2.75--5.12 hours), and a further 2{,}798 attempts failed or were retried, overwhelmingly on transient infrastructure errors. Summing job time gives roughly 5{,}000 job-hours of completed work plus about 600 for failed attempts, or approximately 360{,}000 GPU-hours in total. The campaign spanned 31 days. Each 2.0 iteration issued one research call and one brainstorming call, followed by nominally seven parallel implementation calls, one per selected candidate.

\subsection{Research Loop}
\label{sec:loop}
Each iteration proceeds as: (1) \textbf{context assembly} -- current config and accumulated changes, recent wins/losses with metric deltas, the "What Didn't Work" blocklist, human strategic direction, and (in the research phase) related architecture code and paper summaries; (2) \textbf{hypothesis generation} -- a single coherent modification under strict output contracts; (3) \textbf{implementation} -- one coherent change per candidate to avoid confounded edits, with seven implementation calls running in parallel in MLE agents 2.0; (4) \textbf{training} -- a distributed job (candidates run in parallel in MLE agents 2.0); (5) \textbf{evaluation} -- reading calibrated metrics after job completion (median recorded elapsed time 3.54 hours in the 2.0 campaign); (6) \textbf{decision} -- accept/reject on the variance-aware threshold, with rejects logged to the blocklist.

\begin{figure}[h]
  \centering
  \includegraphics[width=\linewidth]{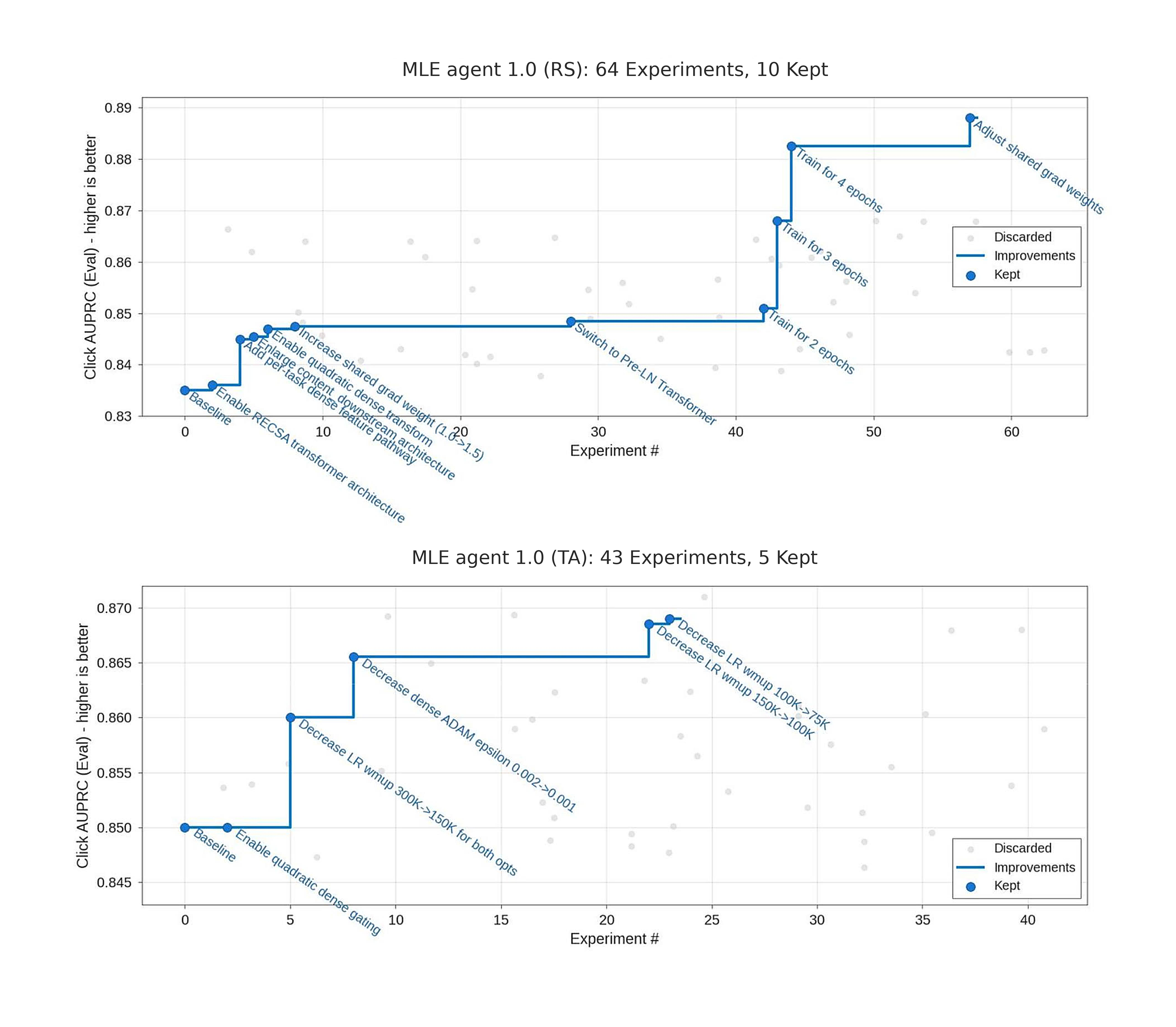}
  
  \caption{\textit{MLE agents 1.0 progress on two surfaces (Click AUPRC, higher is better): Nullstate, 64 experiments with 10 kept; Typeahead, 43 experiments with 5 kept. The same loop compounded improvements on both surfaces.}}
  \label{fig:gen1}
\end{figure}

\textbf{MLE agents 1.0.} We first ran the loop as an exploratory system on two large-scale ranking models, optimizing Click AUPRC (higher is better). On Nullstate it ran 64 experiments and kept 10; on Typeahead, 43 experiments and kept 5 (Figure~\ref{fig:gen1}). That one loop produced compounding improvements on two distinct surfaces was our first evidence that the approach generalizes rather than exploiting a single model's quirks.

\paragraph{What 1.0 taught us.}
Three limitations shaped the redesign. First, the objective was miscalibrated: Click AUPRC over-indexed on trivially-predictable direct-navigation results and did not track online impact, so offline "wins" did not reliably transfer. Second, feeding the agent the full unstructured experiment history caused "prompt bloat," with roughly a third of generated edits malformed. Third, serial execution -- one candidate per iteration -- made the hours-long feedback loop the binding constraint on throughput.

\paragraph{MLE agents 2.0.}
Each learning became a design change: we replaced Click AUPRC with a calibrated weighted-NE objective aligned to online impact; we replaced the full history with structured context -- the "What Didn't Work" blocklist, strict output contracts, a one-coherent-change-per-candidate rule, and an anti-staleness policy (only recent, un-carried winners are fed back); and we ran candidates in parallel (nominally seven candidates plus a variance probe per iteration; actual counts vary with retries and reseeding). Across the 2.0 campaign (Nullstate, 131 iterations), these changes raised the win rate from 15.6\% (MLE agents 1.0 on Nullstate: 10 accepted of 64 experiments) to 29.6\% (MLE agents 2.0 on Nullstate: 325 accepted of 1,097 candidate runs), and cut malformed output from 33.8\% to 0.8\%. Because the objective and search budget also differed, this comparison is not a controlled ablation.

\begin{figure}[h]
  \centering
  \includegraphics[width=\linewidth]{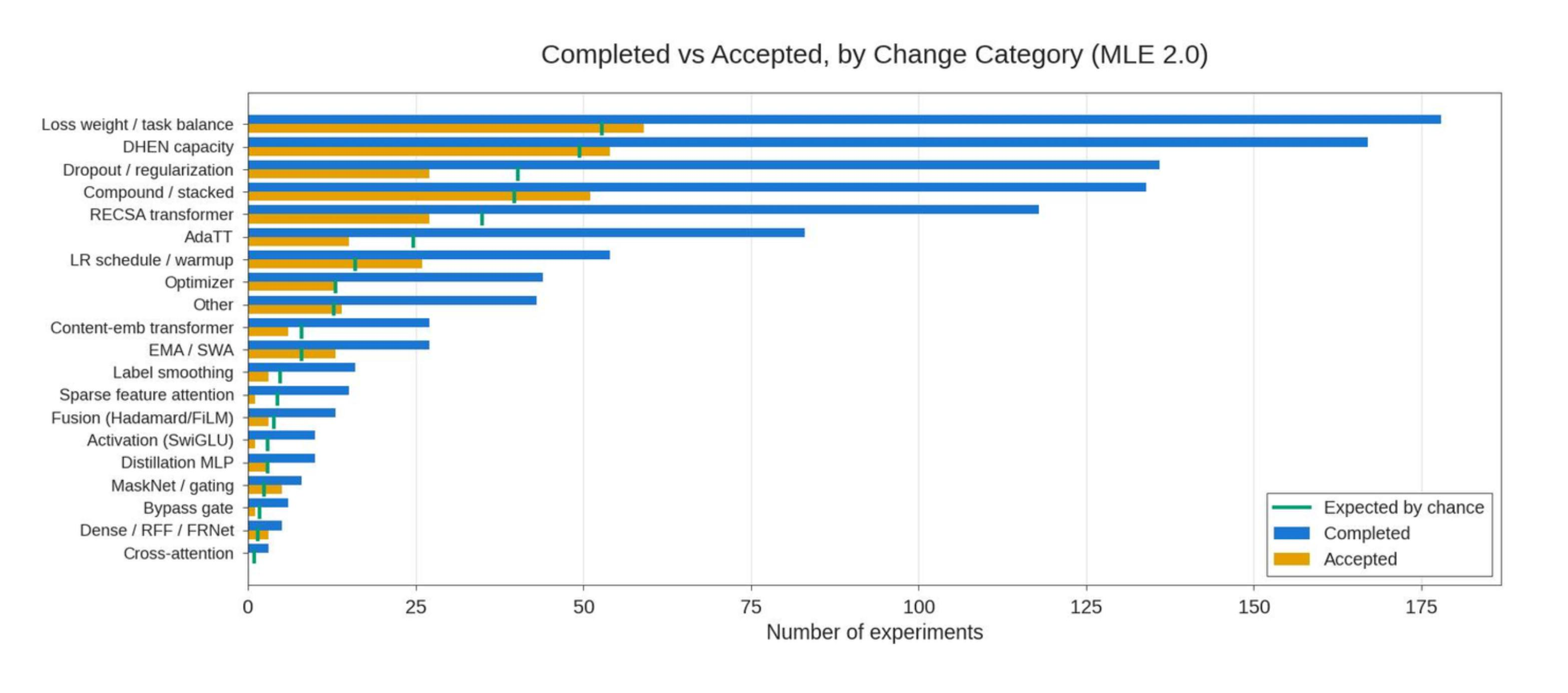}
  \caption{Completed experiments vs.\ accepted, by change category. Orange ticks mark the
  count expected by chance.}
  \label{fig:category}
\end{figure}

\begin{figure}[h]
  \centering
  \includegraphics[width=\linewidth]{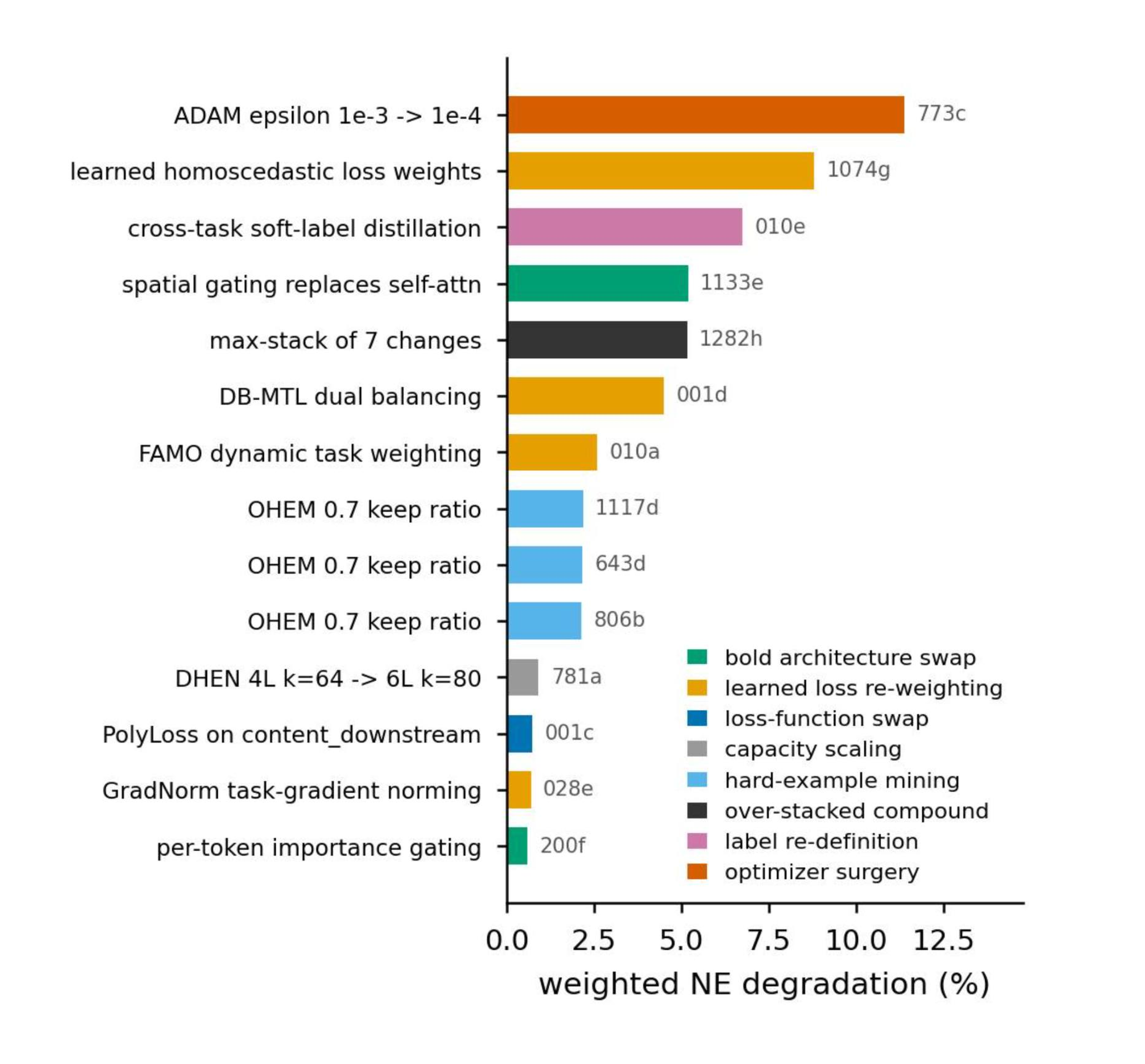}
  
  \caption{Largest logged regressions on the ``What Didn't Work'' blocklist (weighted NE
  degradation vs.\ baseline).}
  \label{fig:blocklist}
\end{figure}

\subsection{Human Interventions}
MLE agents 2.0 runs \textbf{human-on-the-loop}: it launches and evaluates candidates continuously without blocking on a person, while an engineer periodically steers it through the same context channels the agent already reads. Examples of human intervention across development of the two generations were: (i) seeding specific ideas into the seeds file when progress stalled (for example, pointing the agent at content-embedding transformers); (ii) writing failed directions into the "What Didn't Work" blocklist after the agent looped on them -- most notably a 0-for-8 streak on optimizer tuning in 1.0, after which we redirected it toward architecture search; (iii) recalibrating the objective once we diagnosed that click AUPRC was rewarding trivially-predictable direct-navigation results; and (iv) tightening training constraints (e.g., cutting LR warmup from 300K to 200K steps) to keep candidates deployable. None of these paused the loop -- each took effect on the next iteration while the agent kept running.

\subsection{Verification}
\label{sec:verification}
We use verification at three points. \textbf{During the run}, a candidate is accepted only if it beats its same-iteration variance probe by at least 0.0005 NE, screening out improvements smaller than the operational training-variation threshold. Each job's completion is confirmed through two independent status signals, and jobs that fail for transient reasons are automatically resubmitted (up to two more times), so a crashed or mis-logged run is never mistaken for a result. \textbf{Offline}, every candidate is scored on fixed evaluation sets (131,072 click and 55,234 content\_kw\_hc samples) so deltas are comparable, and an engineer reviews the final accumulated stack for coherence offline. \textbf{Online}, the model is evaluated in an independent experiment (+0.23\% search-volume results).
The 325 offline acceptances should therefore be interpreted as search decisions.

\subsection{Significance of result}
The significance is that an autonomous agent's work cleared---and survived---an industry-level
bar far higher than a benchmark's. Prior work evaluates ML-engineering agents on curated
benchmarks and offline task suites~\citep{aiscientist,aide,mlebench,rebench}; here the model serves a large-scale ranker and moved top-line metrics in a controlled experiment (+0.23\% search volume), not a
held-out score. Several of its accepted changes (RFF dense gating~\citep{rahimi2007},
learned-sigmoid attention gating, a zero-init gated distillation bypass) were mechanisms the team had not previously tried, drawn from the broader industry and literature. The agent's contribution was sourcing and composing them for a large-scale ranker where they had not been applied. The transferable lesson for autonomous ML research is that the leverage sits in the loop, not the model: a calibrated objective, structured context, and parallel evaluation turned a general LLM into a productive ML engineer on a live industrial system.

\subsection{Interpretability}
Every model candidate is tracked as a row in a persistent state file: its hypothesis text, the config commit hash, the training-job ID, its four evaluation metrics (click and content NE on the full and keyword+high-confidence slices), the resulting weighted-NE delta, and the accept/reject decision. We also store the running "What Didn't Work" blocklist -- each rejected direction with its measured regression (e.g., ADAM-epsilon change +6.2\% NE, homoscedastic-uncertainty weighting +4.5\%, DB-MTL balancing +1.8\%) -- and mirror the state file to a spreadsheet for monitoring. Because every accepted change is an atomic commit, the model replays step by step from the baseline commit, and any number we report traces back to the specific training run that produced it.

\subsection{Limitations}
The approach has clear limits. \textbf{Exploration collapse:} the agent can spiral within a failing category -- in 1.0 it went 0-for-8 on optimizer tuning before a human redirected it -- and the blocklist mitigates but does not prevent this. \textbf{Unreliable self-correction:} it sometimes prunes its own bad decisions (it reverted a quadratic-gating change after 18 later experiments showed it was net-negative) but not consistently. \textbf{Diminishing returns:} within a single campaign the best model was found well before the end and later iterations added nothing, so more compute on the same search space does not help. \textbf{Offline-online gap:} not every offline winner moves online, and the agent has no mechanism beyond the calibrated objective to predict which will -- nor does it run the online experiments itself.

\section{Broader Impact}
MLE agents do not replace ML engineers -- they change the role from \textit{hypothesis executor} to \textit{research director}. The human contribution shifts from "form a hypothesis, implement it, check the result" to "set the objective, calibrate the reward, diagnose failure modes, and decide what to deploy" -- a higher-leverage role that demands deeper domain expertise, not less. On credit and authorship: when an agent discovers RFF gating, learned attention gating, sparse pooling, and a dozen other changes, we argue the intellectual credit belongs to the system designers who built the loop, calibrated the objective, and structured the context; the agent is a productive tool.

Our experience suggests a flywheel - better agents produce more wins, more wins produce better offline-to-online calibration data, and better calibration produces smarter agents - though returns diminish within a single campaign, so sustaining progress requires expanding the hypothesis space (research phase, new model surfaces, multi-agent teams) rather than merely running more iterations. We believe the community should prepare for a world where autonomous agents routinely contribute to ML research, and that the field should focus on: (1) verifying agent-generated results with the same rigor applied to human research, (2) crediting the human judgment that makes agent research productive, and (3) ensuring autonomous research strengthens rather than dilutes scientific standards. Concretely, we advocate mandatory online/holdout verification of agent claims, durable logging of every experiment and decision, and explicit human accountability for artifacts.

\clearpage
\appendix

\section{Reproducibility Statement}
The primary result is a model-improvement campaign on an industry-scale ranker. To support reproducibility of the \emph{method}, we specify: the calibrated
objective of Equation~\ref{eq:objective} (on the keyword + high-confidence slice), the
0.0005 NE variance-probe win threshold, the six-step research loop of
Section~\ref{sec:loop}, the context-engineering components (blocklist, one-change rule,
strict output contracts, anti-staleness), and the parallel-training regime (seven
candidates plus a variance probe per iteration, across 131 iterations). Compute consisted
of industry-scale distributed-training jobs with 64 GPUs per job, with a median recorded elapsed time of 3.54 hours (mean 4.02; 10th--90th percentile 2.75--5.12 hours) in the 2.0 campaign.
The architecture discoveries in Table~\ref{tab:discoveries} are described in sufficient
detail to be re-implemented on comparable multi-task ranking stacks, and online
verification followed the controlled-experiment protocol in
Section~\ref{sec:verification}.

\section{Disclosure Statement}
\paragraph{Agent(s) / model(s) used}
An LLM-driven MLE agent (Claude Opus 4.8, within the harness of
Section~\ref{sec:loop}) generated the model changes; the human authors designed the
harness, curated the campaign, and verified the work.

\begin{itemize}
  \item[$\boxtimes$] A human author has curated this work and verified every claim,
    figure, methodology detail, and reference.
\end{itemize}

\section{References}
\begingroup
\renewcommand{\section}[2]{}

\endgroup

\section*{NeurIPS Paper Checklist}

\begin{enumerate}

\item {\bf Claims}
    \item[] Question: Do the main claims made in the abstract and introduction accurately reflect the paper's contributions and scope?
    \item[] Answer: \answerYes{}
    \item[] Justification: The abstract and Section~1 state the large-scale system setting, the agent's role in the experimental loop, and the main offline and online results; Sections~3.4 and~3.8 delimit human control and the system's current limitations. 

\item {\bf Limitations}
    \item[] Question: Does the paper discuss the limitations of the work performed by the authors?
    \item[] Answer: \answerYes{}
    \item[] Justification: Section~3.8 discusses exploration collapse, unreliable self-correction, diminishing returns, and the offline--online gap, while Section~3.4 makes the remaining human interventions explicit. 

\item {\bf Theory assumptions and proofs}
    \item[] Question: For each theoretical result, does the paper provide the full set of assumptions and a complete (and correct) proof?
    \item[] Answer: \answerNA{}
    \item[] Justification: The paper reports an empirical large-scale system study and does not claim new theoretical results or provide theorem-style analyses.

    \item {\bf Experimental result reproducibility}
    \item[] Question: Does the paper fully disclose all the information needed to reproduce the main experimental results of the paper to the extent that it affects the main claims and/or conclusions of the paper (regardless of whether the code and data are provided or not)?
    \item[] Answer: \answerNo{}
    \item[] Justification: Sections~1.2 and~3.1--3.5 and Appendix~A describe the objective, acceptance threshold, research loop, context-management mechanisms, and evaluation protocol. 

\item {\bf Open access to data and code}
    \item[] Question: Does the paper provide open access to the data and code, with sufficient instructions to faithfully reproduce the main experimental results, as described in supplemental material?
    \item[] Answer: \answerNo{}
    \item[] Justification: 
    Appendix~A documents the reproducible method components that can be described without releasing proprietary assets. 

\item {\bf Experimental setting/details}
    \item[] Question: Does the paper specify all the training and test details (e.g., data splits, hyperparameters, how they were chosen, type of optimizer) necessary to understand the results?
    \item[] Answer: \answerNo{}
    \item[] Justification: Sections~1.2, 3.2, and~3.5 report the data scale, fixed evaluation-set sizes, objective, acceptance rule, training duration, and distributed-training setup. Some implementation details, including the complete data-split construction, optimizer configuration, and full model hyperparameters, are not disclosed.

\item {\bf Experiment statistical significance}
    \item[] Question: Does the paper report error bars suitably and correctly defined or other appropriate information about the statistical significance of the experiments?
    \item[] Answer: \answerNo{}
    \item[] Justification: Table~2 reports 95\% confidence intervals computed using Fieller's method for the primary online A/B metrics, and Section~3.5 describes the variance-probe threshold used offline. Offline uncertainty is not fully quantified. 

\item {\bf Experiments compute resources}
    \item[] Question: For each experiment, does the paper provide sufficient information on the computer resources (type of compute workers, memory, time of execution) needed to reproduce the experiments?
    \item[] Answer: \answerYes{}
    \item[] Justification: Section~3.2 and Appendix~A report distributed jobs using 64 GPUs per job, with a median recorded elapsed time of 3.54 hours and a mean of 4.02 hours in the 2.0 campaign. 
    
\item {\bf Code of ethics}
    \item[] Question: Does the research conducted in the paper conform, in every respect, with the NeurIPS Code of Ethics \url{https://neurips.cc/public/EthicsGuidelines}?
    \item[] Answer: \answerYes{}
    \item[] Justification: The work followed the NeurIPS Code of Ethics. 

\item {\bf Broader impacts}
    \item[] Question: Does the paper discuss both potential positive societal impacts and negative societal impacts of the work performed?
    \item[] Answer: \answerNo{}
    \item[] Justification: Section~4 discusses potential benefits for ML-engineering productivity, scientific verification, and human accountability. It does not yet give a comparably concrete discussion of risks such as reward misspecification, adaptive overfitting, unsafe model changes, increased compute use, or disparate effects across users. 
    
\item {\bf Safeguards}
    \item[] Question: Does the paper describe safeguards that have been put in place for responsible release of data or models that have a high risk for misuse (e.g., pre-trained language models, image generators, or scraped datasets)?
    \item[] Answer: \answerNA{}
    \item[] Justification: The work does not release a dataset, pretrained model, or other high-risk asset. 

\item {\bf Licenses for existing assets}
    \item[] Question: Are the creators or original owners of assets (e.g., code, data, models), used in the paper, properly credited and are the license and terms of use explicitly mentioned and properly respected?
    \item[] Answer: \answerYes{}
    \item[] Justification: External methods and models are credited, and no third-party code or data is redistributed. 

\item {\bf New assets}
    \item[] Question: Are new assets introduced in the paper well documented and is the documentation provided alongside the assets?
    \item[] Answer: \answerNA{}
    \item[] Justification: The submission does not release a new public dataset, codebase, model checkpoint, or benchmark artifact. 

\item {\bf Crowdsourcing and research with human subjects}
    \item[] Question: For crowdsourcing experiments and research with human subjects, does the paper include the full text of instructions given to participants and screenshots, if applicable, as well as details about compensation (if any)? 
    \item[] Answer: \answerNA{}
    \item[] Justification: The work involved no crowdsourcing, participant recruitment, researcher-provided instructions, or compensation.

\item {\bf Institutional review board (IRB) approvals or equivalent for research with human subjects}
    \item[] Question: Does the paper describe potential risks incurred by study participants, whether such risks were disclosed to the subjects, and whether Institutional Review Board (IRB) approvals (or an equivalent approval/review based on the requirements of your country or institution) were obtained?
    \item[] Answer: \answerNA{}
    \item[] Justification: The work was not classified as human-subjects research and therefore did not require IRB approval or an institutional equivalent. 

\item {\bf Declaration of LLM usage}
    \item[] Question: Does the paper describe the usage of LLMs if it is an important, original, or non-standard component of the core methods in this research? Note that if the LLM is used only for writing, editing, or formatting purposes and does \emph{not} impact the core methodology, scientific rigor, or originality of the research, declaration is not required.
    \item[] Answer: \answerYes{}
    \item[] Justification: Sections~3.1 and~3.3 describe the controller LLM's role in architecture research, hypothesis generation, and implementation, and Appendix~B identifies the model and separates agent-generated changes from human design, and steering.

\end{enumerate}


\begin{thebibliography}{99}
\bibitem[Rahimi and Recht(2007)]{rahimi2007} Rahimi, A. and Recht, B. Random Features for Large-Scale Kernel Machines. \emph{NeurIPS}, 2007.
\bibitem[Tang et al.(2020)]{ple2020} Tang, H. et al. Progressive Layered Extraction (PLE): A Novel Multi-Task Learning Model. \emph{RecSys}, 2020.
\bibitem[Ma et al.(2018)]{mmoe2018} Ma, J. et al. Modeling Task Relationships in Multi-task Learning with Multi-gate Mixture-of-Experts (MMoE). \emph{KDD}, 2018.
\bibitem[Wang et al.(2021a)]{dcnv2} Wang, R. et al. DCN V2: Improved Deep \& Cross Network for Feature Interactions. \emph{WWW}, 2021.
\bibitem[Mao et al.(2023)]{finalmlp} Mao, K. et al. FinalMLP: An Enhanced Two-Stream MLP Model for CTR Prediction. \emph{AAAI}, 2023.
\bibitem[Wang et al.(2023)]{gdcn} Wang, F. et al. Gated Deep Cross Network (GDCN). 2023.
\bibitem[Wang et al.(2021b)]{masknet} Wang, Z. et al. MaskNet: Introducing Feature-Wise Multiplication to CTR Ranking. 2021.
\bibitem[Lu et al.(2024)]{aiscientist} Lu, C. et al. The AI Scientist: Towards Fully Automated Open-Ended Scientific Discovery. 2024.
\bibitem[Jiang et al.(2025)]{aide} Jiang, Z. et al. AIDE: AI-Driven Exploration for ML Engineering. 2025.
\bibitem[Chan et al.(2024)]{mlebench} Chan, J. et al. MLE-bench: Evaluating Machine Learning Agents on ML Engineering. 2024.
\bibitem[Wijk et al.(2024)]{rebench} Wijk, H. et al. RE-Bench: Evaluating Frontier AI R\&D Capabilities of Language Model Agents. 2024.
\bibitem[Wang et al.(2026)]{wang2026selfevolving}
Wang, H., Wu, Y., Chang, D., Wei, L., and Heldt, L.
Self-Evolving Recommendation System: End-To-End Autonomous Model
Optimization With LLM Agents.

\emph{arXiv preprint arXiv:2602.10226}, 2026.
\end{thebibliography}
\end{document}